\documentclass[11pt, oneside]{article}   	
\usepackage{geometry}                		
\usepackage{graphicx}				
\usepackage{authblk}
\usepackage{times}
\usepackage{subfigure} 
\usepackage{wrapfig}
\usepackage[numbers]{natbib}
\usepackage{algorithm}
\usepackage{algorithmic}
\usepackage{bm}
\usepackage{amssymb}
\usepackage{amsmath}
\usepackage{graphicx}

\title{Forecasting Crime with Deep Learning}
\author[1]{Alexander Stec \thanks{stec@u.northwestern.edu}}
\author[2]{Diego Klabjan}
\affil[1]{Engineering Sciences and Applied Mathematics, Northwestern University}
\affil[2]{Industrial Engineering and Management Sciences, Northwestern University}

\date{}							

\begin{document}
\maketitle
\begin{abstract} 
The objective of this work is to take advantage of deep neural networks in order to make next day crime count predictions in a fine-grain city partition.
We make predictions using Chicago and Portland crime data, which is augmented with additional datasets covering weather, census data, and public transportation.
The crime counts are broken into 10 bins and our model predicts the most likely bin for a each spatial region at a daily level.
We train this data using increasingly complex neural network structures, including variations that are suited to the spatial and temporal aspects of the crime prediction problem.
With our best model we are able to predict the correct bin for overall crime count with 75.6\% and 65.3\% accuracy for Chicago and Portland, respectively.
The results show the efficacy of neural networks for the prediction problem and the value of using external datasets in addition to standard crime data.
\end{abstract} 

\section{Introduction}\label{sec:introduction}
\label{submission}
Police departments have spent a lot of time and resources to discover crime rate trends and how to adapt their policing techniques along with these trends.
Accurately predicting trends has obvious benefits, such as properly assigning patrols, and it allows police departments to be proactive rather than reactive.
At first, police were limited to looking at historical trends and extrapolating for the future, but as analytics-based tools have improved, models with much more predictive power began to emerge.
Of particular interest are the attempts using machine learning techniques, which were introduced as early as 1998, but were limited by computational power at that time.

Before machine learning, the first techniques simply extrapolated historical trends.
This works well at very long time scales, but obviously cannot predict deviations from these trends.
Slightly more advanced techniques, such as multivariate regression, did improve predictive power, but still only relied on historical crime data.
There are other factors that are known to affect crime rates, such as weather and socioeconomic factors \cite{cohn1990weather,carlen1988women}, and the mentioned methods do not take these into account.
The use of additional datasets do cause the number of features to rise greatly, and newer machine learning methods are better suited to handle a large number of features and find the complex relationships between these factors.

With the development of high-performance computers, the cost of these machine learning techniques was drastically reduced.
They achieved a great deal of success in many domains.
In particular, deep learning, a relatively recent development in artificial intelligence, has achieved impressive results with many types of classification problems, ranging from speech to visual recognition.
One domain that has not received much attention within deep learning is crime prediction.
Deep learning is well suited to handle the temporal and spatial components of the problem as well as a large feature set that can be constructed using relevant public datasets in addition to crime datasets.
In addition, the inherent complexity of the problem makes it suitable for deep learning.
Using neural networks reduces the need for extensive feature engineering found in some previous work and allows for training over large datasets.
In order to address the temporal aspects of crime prediction, a Recurrent Neural Network (RNN) can be used. 
In a traditional neural network, there is the assumption that all inputs and outputs are independent of one another, but for crime this is clearly not the case.
Crimes that happened yesterday will affect what happens today.
For example, if there is an increase in thefts in an area then it is possible that the area is being targeted and will continue to have theft rates higher than normal for a period of time.
RNNs apply the same weights to each element of a sequence while also having a memory of past states of the sequence. 
This memory encodes the sequence up to the given point in time (it is also called the hidden state).
This means a given output at a time depends on the current input at the time and the hidden state containing the history up until the time, allowing for greater predictive power.

For the spatial aspects of crime prediction, a Convolution Neural Network (CNN) is fitted.
CNNs are widely used and successful in image classification.
For this task, the image is input as a grid of pixels with three channels, corresponding to RGB values.
For our purposes, we can think of a city grid with relevant data in each grid in terms of an image. 
Each grid cell corresponds to a pixel, and each different feature in that cell corresponds to a different channel.
CNNs slide filters across an image to get local responses from the image, and then use a max pooling over these responses to reduce the dimensionality of the problem.
As dimensionality is reduced, the features extracted go from highly local to a global view built off the aggregate of the local features.
Since crime is not randomly distributed geographically, taking into account what is happening in neighboring cells is useful for predictions in the main cell.

While RNNs and CNNs capture temporal and spatial aspects of the problem separately, they can be combined into a single network that captures both.
Network architecture is flexible and so it is possible to include both types of layers into one network.
The most important spatial features at a time are placed into a grid and passed through the CNN layers, and the output of the CNN is combined with the rest of the features.
The resulting vector is then treated as input at the time for the RNN.

The common dataset found in all work on the topic is some sort of a crime dataset that includes geographic and temporal information.
The crime data can be anything from a reported crime to a 911 emergency call.
For our purposes, the crime dataset contains precise latitude and longitude of crimes, detailed crime type classification and codes, and additional information such as whether or not an arrest was made and a flag for domestic crimes.
However, there has not been much utilization of outside datasets.
There are many publicly available datasets that can augment the traditional crime data.
We make use of weather data, census data, and public transportation data on top of the crime reports.
Cohn \cite{cohn1990weather} establishes a connection between weather and crime.
Additionally, although the crime information is highly granular, even down to the minute in many cases, previous work was limited to predicting aggregate trends over longer time intervals of a month or week.
While this is helpful, increasing the granularity of the predictions is clearly desirable, though it is limited by the amount of crime occurrences in a given area.
We predict at the daily-beat level, because it is at this level that our additional datasets are the most relevant.

This work contains three main contributions.
First, we are the first to use a joint recurrent and convolutional neural network for the purpose of predicting crime.
As previously mentioned, these architectures are well suited to handling both the temporal and spatial aspects of crime patterns.
Second, we are the first to combine crime data with additional weather, public transportation, and census data.
Crime trends are the result of complex interactions between various environmental, economic, and sociological situations, and so including this type of data on top of raw crime data allows for more insight.
Third, we are the first work to make predictions for the next day.
While predictions at the monthly and weekly level do serve a purpose, having predictions at the daily level would allow police to be much more precise with their efforts.

\section{Related work}
Methods used for crime prediction have changed greatly over the years as new analytics-based software was introduced. 
Geographical Information Systems (GIS) were the first and are the most widespread analytic tool for spatial data \cite{groff2002forecasting}.
GIS is quite useful at producing maps and retrospectively finding links between crime patterns and various spatial and social conditions \cite{longley2005geographic,ormsby1999extending}, but does not provide much predictive power on its own.

One method relying mostly on GIS is the use of crime hot spots, which simply assumes that past crime patterns will predict future crime patterns.
The most common method was introduced by Eck et al. \cite{eck2005mapping}, and uses kernel density estimation.
This method did work well over long periods of time, but it was found that even on a monthly scale it is a poor predictor \cite{adams2001historical,jefferis1999multi}.
Since it uses only crime data, it also is poorly suited to pick up new trends before they are seen.

Better results were obtained using multivariate regression and a careful selection of indicator variables, as seen in the works by Gorr et al. \cite{gorr2000assessment,gorr2002crime}.
Gorr et al. found that the success of this method relied heavily on properly choosing the indicator variables, but with good selection it performed far better than the hot spots method.
Unlike the hot spots method, using multivariate regression allows for forecasts that differ greatly from historical variation.
The limit of this method is determined by the number of data points, as there have to be enough events in each grid to predict accurately.
At the monthly temporal level, Gorr et al. found that they could use a 4000 x 4000 square feet grid to get acceptable errors with their Pittsburgh crime data.

First attempts at using neural networks did find that they are able to outperform regression methods, but they were limited by the computational power at the time \cite{olligschlaeger1997artificial}.
Current advances in hardware and in network modeling make neural networks more appealing and powerful.
One paper similar to \cite{olligschlaeger1997artificial} that incorporated a simple shallow feed forward neural network, Bogomolov et al. \cite{bogomolov2014once}, attempted to predict crime surges in London areas using mobile phone, crime, and census data.
The time scale used was again monthly, although the spatial grids were small, incorporating a population of about 1,500 people.
There is a disparity between the time scale of the mobile phone data (hourly) and the prediction scale (monthly), but various aggregation methods for the mobile data were used.

The work of Yu et al. \cite{yu2011crime} used a simple shallow feed forward network and tried to predict burglary type crime hot spots at the monthly level.
The dataset consisted of crime records with location as well as crime related events, though they did not specify the size of their grid cells.
Additional features were obtained by leveraging temporal knowledge by using previous month's data to predict the future month, and also creating local features by averaging these counts over neighboring grid cells.

With regards to input, the common thread found in all previous work was detailed location tagged crime data, e.g. crime records or 911 calls.
On top of this, some works included additional datasets, but they are not found consistently. 
Our work utilizes census data that has been seen before and other data that is new (daily weather reports and public transportation information).
This information in some major cities is readily available through public datasets and is an excellent augmentation to the standard crime data found in all previous work.

%
Crime as a problem of course has both spatial and temporal dimensions, and some components of these were not accounted for in previous work.
In the work by Yu et al. \cite{yu2011crime}, the spatial dependence was captured through feature engineering, and many works have used time series for temporal dependence \cite{yu2011crime,greenberg2001time}.
This type of information can be covered better by certain network architectures; recurrent networks are used to handle temporal information and convolutional networks for local spatial information.
Our main model does utilize both convolution and recurrent layers to capture the temporal and spatial aspects of the data.

\section{Models}
In this work we use four different types of neural networks to forecast crime, which are described in this section in order of increased complexity.
The specifics of data preparation are discussed in Section 4, but the basic format is as follows for the purpose of this section.
Each city is split into grid cells (beats for Chicago and a square grid for Portland), and within each cell there are a set of features corresponding to a certain day.
One record contains all the features of all cells for a given day.
The prediction of crime count for the next day is done by predicting one of some number of bins, $B$, for the count in each cell (e.g. crime count in ranges 0-5, 6-10, etc.).

All of our networks follow a supervised learning approach, meaning we have knowledge of the correct bins for each record.
Mathematical formulations of the networks and training process are included in the following subsections, but at a high level the training process is as follows.

\begin{enumerate}
\item Initialize network weights randomly
\item For each record in data
\begin{itemize}
\item Input record into network
\item Perform forward pass with current weights to get predicted output
\item Compute the loss function that depends on the distance between predicted output and true output
\item Calculate gradient by taking derivative of loss function and backpropagation
\end{itemize}
\item Average gradient over all records
\item Update weights using the gradient multiplied by a learning rate
\item Repeat steps 2-4 until network training stabilizes
\end{enumerate}

After enough iterations of this process, the network improves from random outputs to accurately matching true outputs.
An important advantage of using deep neural networks is that there is no need to select important features, because the network is able to determine which features have the most predictive power on its own.
It is for this reason that our model can handle a much larger number of features compared to previous methods.

\subsection{Feed forward network}
The first model we use is the simplest, a feed forward network. 
This network consists of several layers of units, where the first layer is the input, the last layer is the output, and the inner layers are called hidden layers.
The input and output layers have fixed sizes.
The size of the input layer corresponds to the number of features in the data, and the output layer has the same number of units as the number of possible predictions.
The hidden layers have no connection to the data or prediction, hence the name, and so the number of hidden layers and their sizes are determined experimentally.
Increasing the number of hidden units increases the network's abilities to learn, but too many hidden units tend to overfit.

Each unit is connected to every unit in the previous and following layer by a weight, see Figure 1.
Each data record is inserted into the input layer, and then using the network weights, the values are fed forward through the network until the output layer.
The values are fed from one layer to another by performing a matrix multiplication (which includes a bias) and then putting the results of this through a nonlinear activation function.
In our work, we use rectified linear units for the hidden layers and a softmax activation for the final layer.
The softmax is key for our predictions, as it gives the probability of each classification value.

Mathematically, the hidden layers behave in the following manner\begin{equation}
x_{t, l+1} = f\left( x_{t, l} ; W_{l}, b_{l} \right), \hspace{4 mm} l = 0,...,L \nonumber
\end{equation}
where $x_{t, l}$ is the input to layer $l+1$ at time $t$ and $f$ is a nonlinear function representing the neural network which is parametrized by matrix $W_{l}$ representing the weights for the layer and by the bias vector $b_{l}$.
The first layer is the input layer and it is where our model's features $x_{t, 0}$ enter the network.
We drop the superscript and refer to the model input as $x_{t}$.
We use a rectified linear activation, meaning $f = \max\left(0,s\right)$ with $s = Wx + b$.

Let $C$ be the number of spatial units (cells or beats).
For a network with $L$ hidden layers, the softmax layer takes the final activation $W_{L}x_{t, L} + b_{L} = z_{t} = \left( z_{t1}, z_{t2}, ..., z_{tC} \right) \in \mathbb{R}^{B \cdot C}$, where $z_{tc} \in \mathbb{R}^{B}$ for $c = 1,...,C$, as input, and outputs a vector of probabilities $\bf{o}$ for each bin, where the probability of bin $j \in \{ 1,...,B \}$ is\begin{equation}
o_{tc}^{j}\left( x_{t} \right) = \frac{\exp{\left(z_{tc}^{j}\right)}}{\sum\limits_{j=1}^{B}{\exp{\left(z_{tc}^{j}\right)}}}. \nonumber
\end{equation}
For one softmax layer, the loss is given by\begin{equation}
L_{tc} = \sum\limits_{j=1}^{B}{y_{tc}^{j}\log{o_{tc}^{j} \left( x_{t} \right)}}, \nonumber
\end{equation}
where $y_{tc}^{j} = 1$ if the crime count corresponding to record $x_{t}$ in cell $c$ falls in bin $j$ and $0$ otherwise. 
Of course, we fit the model for each beat and input record, and so our loss function is given by\begin{equation}
L = \sum\limits_{t} \sum\limits_{c=1}^{C} L_{tc} .
\end{equation}

The loss is back-propagated through the network by taking derivatives of the activation functions and the matrix multiplication by use of the chain rule.
Using loss $L_{tc}$ for a single record and starting at the softmax layer, we calculate\begin{equation}
\frac{\partial L_{tc}}{\partial W_{L}^{ij}} =  \frac{\partial L_{tc}}{\partial z_{tc}^{j}} \frac{\partial z_{tc}^{j}}{\partial W_{L}^{ij}}. 
\end{equation}
This is derivative is calculated at all times $t$ and cells $c$ and then averaged to update the weights.
Expanding, we get\begin{equation}
\frac{\partial L_{tc}}{\partial z_{tc}^{j}}  = \sum\limits^{B}_{k=1} \frac{\partial L_{tc}}{\partial o_{tc}^{k}} \frac{\partial o_{tc}^{k}}{\partial z_{tc}^{j}} \nonumber
\end{equation}
\begin{equation}
\frac{\partial L_{tc}}{\partial o_{tc}^{j}} = \frac{y_{tc}^{j}}{o_{tc}^{j}} \nonumber
\end{equation}
\begin{equation}
\frac{\partial o_{tc}^{k}}{\partial z_{tc}^{j}}  = 
\begin{cases}
o_{tc}^{j}\left( 1 - o_{tc}^{j} \right) & \text{if}\ j=k \\
-o_{tc}^{j}o_{tc}^{k} & \text{if}\ j\ne k
\end{cases}
\nonumber
\end{equation}
\begin{equation}
\frac{\partial z_{tc}^{j}}{\partial W_{L}^{ij}} = x^{i}_{t,L}. \nonumber
\end{equation}
Performing the summations and combining the above terms, we simplify (2) and bring back the subscripts to find the gradients at the final layer are given by\begin{equation}
\frac{\partial L_{tc}}{\partial W_{L}^{ij}} = (o_{tc}^{j}-y_{tc}^{j})x_{t,L}^{i}. \nonumber
\end{equation}
To find the gradients for the lower layer weights, we start with
\begin{equation}
\frac{\partial L_{tc}}{\partial W_{l}^{ij}} =  \frac{\partial L_{tc}}{\partial s_{t}^{j}} \frac{\partial s_{t}^{j}}{\partial W_{l}^{ij}}, \nonumber
\end{equation}
where $s_{t,l} = W_{l}x_{t,l} + b_{l}$, and apply the chain rule to the first term for all activations from the final layer to layer $l$.

Once the gradients are calculated, the weights are updated according to the equation\begin{equation}
W_{l}^{ij} = W_{l}^{ij} - \alpha \sum\limits_{t} \sum\limits_{c=1}^{C} \frac{\partial L_{tc}}{\partial W_{l}^{ij}}, \nonumber
\end{equation}

These equations are applicable to all subsequent models.
During training, it is most efficient to split the training records into minibatches and then to calculate the gradient for each record in parallel.
After the gradients are all computed, the average is taken and the network weights are updated before the next minibatch is introduced.

\subsection{Convolutional network}
Convolutional networks were built to handle images, but for the purposes of our problem, we can think of a map of a city as an ``image."
In this ``image," each grid cell corresponds to a pixel and the value of each different feature corresponds to the pixel value for that feature's channel.
A convolutional network works by first sliding a number of smaller filters over an image in the convolutional layers.
This computes a dot product between the weights in each filter and the local region of the image that the filter is currently covering.
The forward pass using each of these filters is the same as in the feed forward network, except now only a subset of the neurons are being used for calculating the input to any given neuron in the next layer.
With a $k \times k$ filter, a neuron $z_{ij}$ in a convolutional layer is calculated by
\begin{eqnarray}
s_{ij} &=& \sum\limits_{d=1}^{D} \left( b^{d} + \sum\limits_{n=0}^{k-1}\sum\limits_{m=0}^{k-1}W^{d}_{nm} x^{d}_{(i+n)(j+m)} \right)\nonumber \\
z_{ij} &=& {\rm relu}(s_{ij}) = \max \left(0,s_{ij} \right), \nonumber
\end{eqnarray}
where $x_{ij}$ is in the input grid and $d = 1,...,D$ are the channels of the input.
This calculation is done for each neuron in each channel in the convolutional layer.

After an activation function is applied to the dot product, the resulting pixel grid is downsampled in the spatial dimensions by means of max pooling.
Max pooling works by sliding a smaller grid, of spatial extent $F$ and stride $S$, over the pixels in each channel and only keeping the value of the highest responding pixel in that grid.
For an $N \times N$ grid with $D$ channels, the output grid is of size $(N-F/S)+1 \times (N-F/S)+1$ with the same number of channels $D$.

This sequence of convolutions and max pooling is performed some number of times until the final pixel grid is flattened, and then used as input to a traditional feed forward network for classification.
Since the weights of the filters are fixed as they pass over the pixels, the filters in the lower levels learn local features seen at a small spatial scale.
As the pixels are downsampled, the later filters pick up on global features built from the lower level ones.
A diagram of a convolutional network is presented in Figure 2, and an example of max pooling can be seen in Figure 3.

\subsection{Recurrent network}
In a traditional neural network, it is assumed that all input and output pairs are independent from one another, but for a time dependent problem like crime this is a very questionable option.
It is possible to take into account time by including features from previous records on top of the current one, but even then the model can only look a fixed window into the past from every record.

Recurrent networks address this shortcoming by looking at sequences of input, and sharing the feed forward weights between all these inputs.
Additionally, the hidden layers of the each input are connected to the hidden layers of the input immediately before and after.
The hidden layers from the previous input are called the hidden state of the network, and it is essentially a memory that captures what happened at the previous time steps.
In a sequence, the first input is run through the network, and for the next step both the next input and current state of the network are used in calculating the output, as seen in Figure 4.
This means that past information can have a strong influence on the prediction for the current input.

Since the previous state of the network is used in the network, the new mapping of a single layer network reads\begin{eqnarray}
z_{t} = f\left( x_{t} ,h_{t-1} ;W,b\right) \nonumber \\
h_{t} = g\left( x_{t,} , h_{t-1} ; W,b \right) \nonumber
\end{eqnarray}
where $h_{t-1}$ is the state of the network at the previous time step.
The network uses the same loss function (1).
In the feed forward model, input records $x_{t}$ can be unordered in time, here, however, $t$ must correspond to time.

In particular, we use a type of RNN called LSTM (Long Short Term Memory network) that is capable of learning long term dependencies.
This is a very popular variant for an RNN cell, and it is widely used in many contexts with a lot of success.

\subsection{Recurrent convolutional network}
We use a standalone recurrent network as one of our models, and also combine recurrent layers with convolutional layers for our final model, which can be found in Figure 5.
For this network, the top spatially correlated features are run through the convolutional network, as outlined in Figure 2, until the final feed forward layer.
At this point, this layer is concatenated with the rest of the features for that day.
The concatenation is the input for the RNN, which behaves in the same way as outlined in Section 2.3.

\begin{figure}[ht]
\begin{center}
\includegraphics[width=10cm]{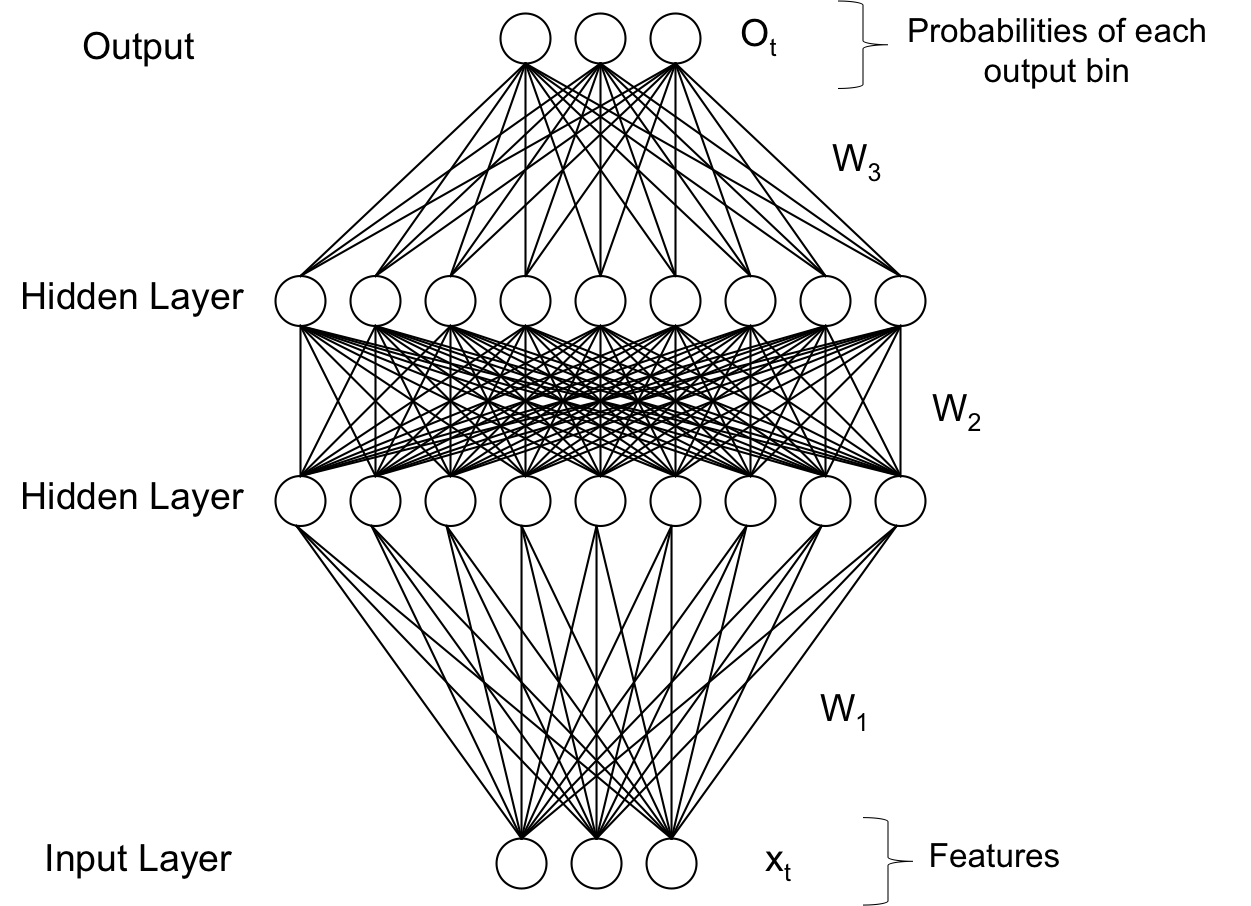}
\caption{Network structure for basic feed forward network.}\label{fig:decoupled}
\end{center}
\end{figure}

\begin{figure}[ht]
\begin{center}
\includegraphics[width=10cm]{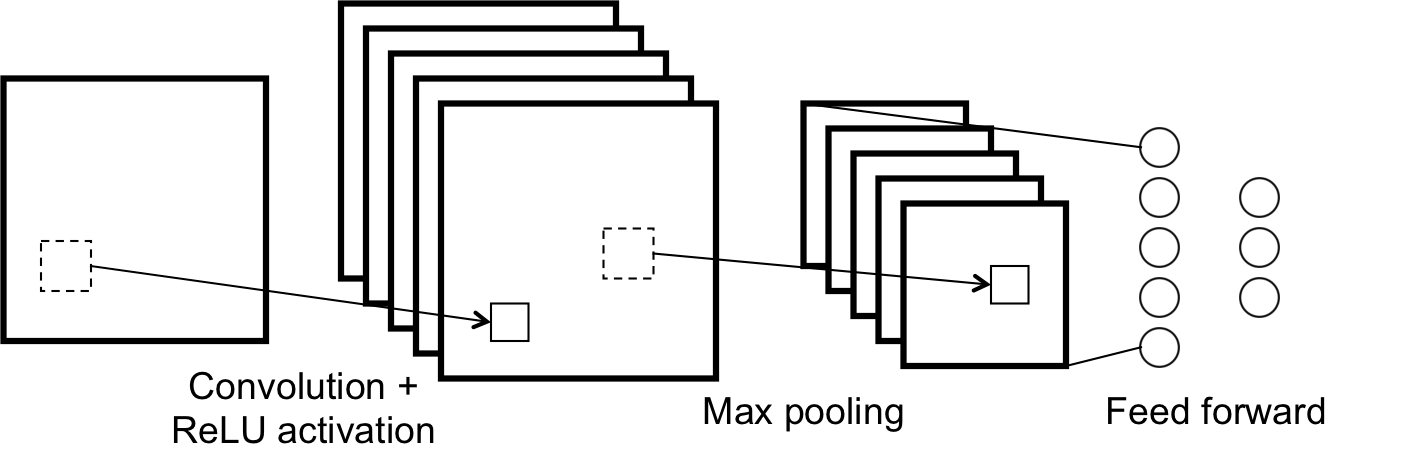}
\caption{Network structure for convolutional network with just one convolution and one max pooling.}\label{fig:decoupled}
\end{center}
\end{figure}

\begin{figure}[ht]
\begin{center}
\includegraphics[width=10cm]{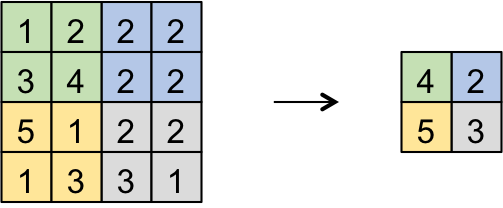}
\caption{Example of maxpooling for a $4\times 4$ grid with a $2\times 2$ filter and stride of 2.}\label{fig:decoupled}
\end{center}
\end{figure}

\begin{figure}[hb]
\begin{center}
\includegraphics[width=12cm]{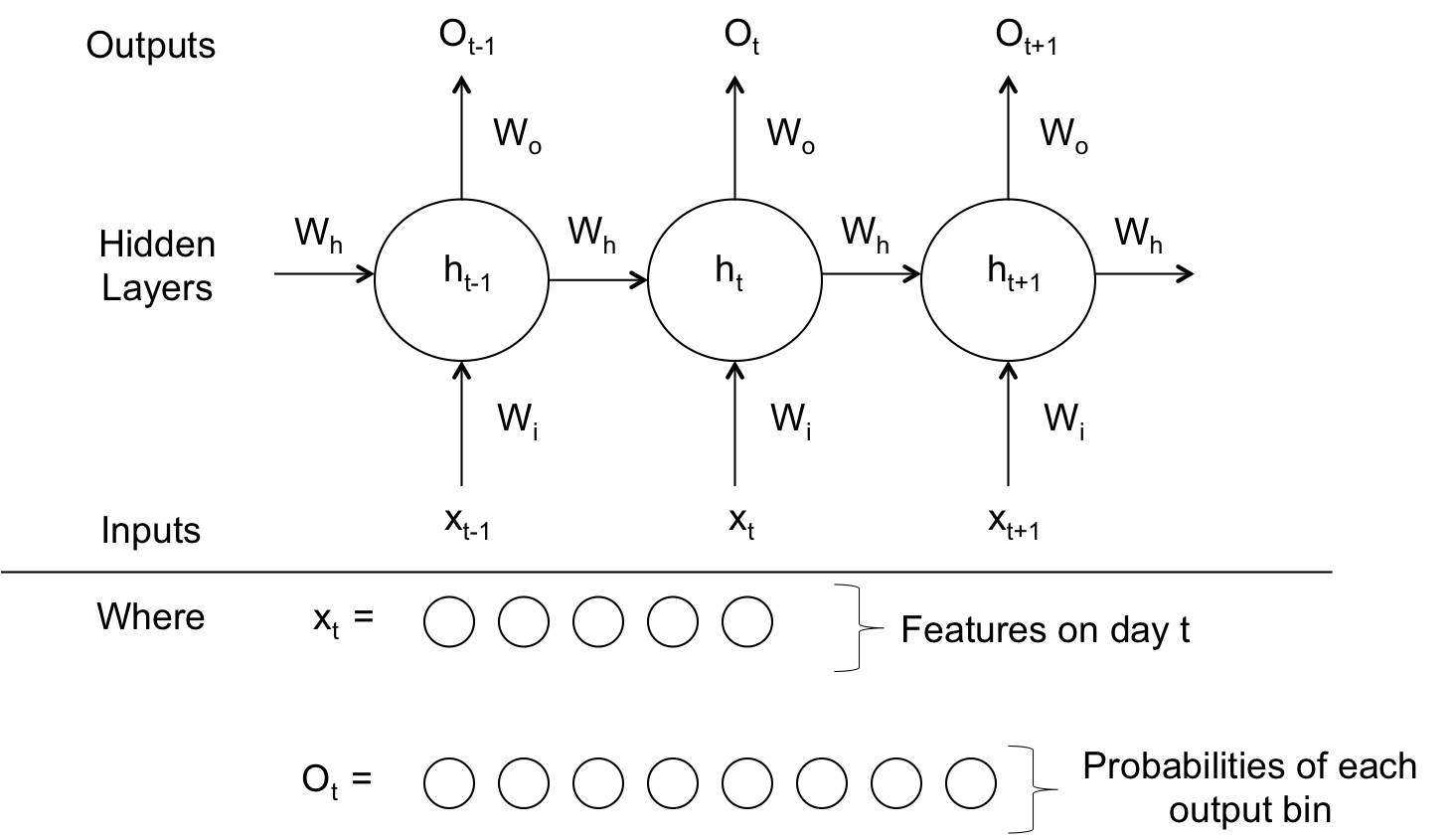}
\caption{Network structure for recurrent network.}\label{fig:decoupled}
\end{center}
\end{figure}

\begin{figure}[hb]
\begin{center}
\includegraphics[width=12cm]{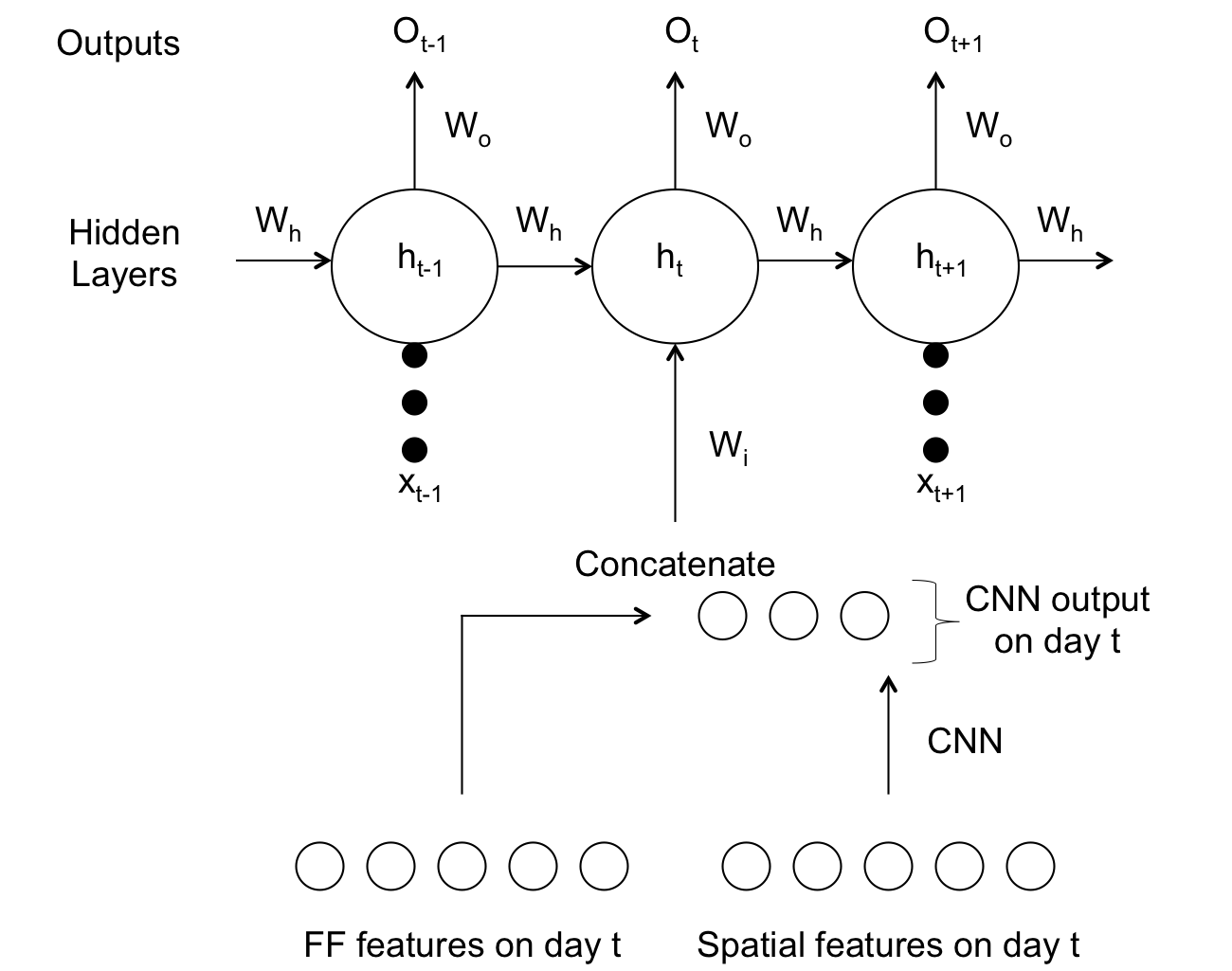}
\caption{Network structure for recurrent convolutional network.}\label{fig:decoupled}
\end{center}
\end{figure}

\clearpage
\section{Model specifics}
\subsection{Data and preparation}
The datasets we use are in the areas of crime, public transportation, weather, and census data.
For Chicago, the first two are publicly available through the city of Chicago's data portal website, https://data.cityofchicago.org/.
For Portland, the crime data was obtained from the National Institution of Justice Real-Time Crime Forecasting challenge.
Unfortunately, there was no source for public transport data in Portland.
The latter two datasets are also publicly available as well.
The census data is available through the United States Census Bureau at https://www.census.gov/data.html, and the weather data is available through the National Oceanic and Atmospheric Administration at https://www.ncdc.noaa.gov/.
The details of each dataset are discussed next.

We split each city in grids, and so all this data must be appropriately mapped to the grids.
For Chicago, we have 274 grid cells, corresponding to the 274 police beats in the city.
The city provides a shapefile of the beats, and so all the datasets that include latitude and longitude can easily be mapped into the correct police beat.
For Portland, only the districts are provided, and these are too large to be useful for predictions.
Instead, we create our own uniform grid with given maximum/minimum latitudes and longitudes. 
We do not use the maximum/minimum of the city limits, since this includes many irrelevant areas such as rivers or the airport, but rather manually choose these values.

For the crime specific data for Chicago, we use roughly 6 million records going back to 2001, were each record is a reported crime with many informative fields.
Of course, geographic information is provided, such as longitude, latitude, beat, district, etc., but there are also fields regarding the type of crime and binary fields, such as if an arrest was made or if it was a domestic crime.
These records are all first aggregated by beat and day to form the input for our networks.
This means the input to the networks corresponds to a given day, and the true labels are the counts of reported crimes for each beat.
We place the true counts into one of ten bins and we use these bins for classification.
Figure 6 shows the counts of each bin over all days for three Chicago beats representative of beats with low, medium, and high standard deviations for crime count.
There is no significant change in the distributions over the timeframe of the dataset.
\begin{figure}[ht]
\begin{center}
\includegraphics[width=1.94in]{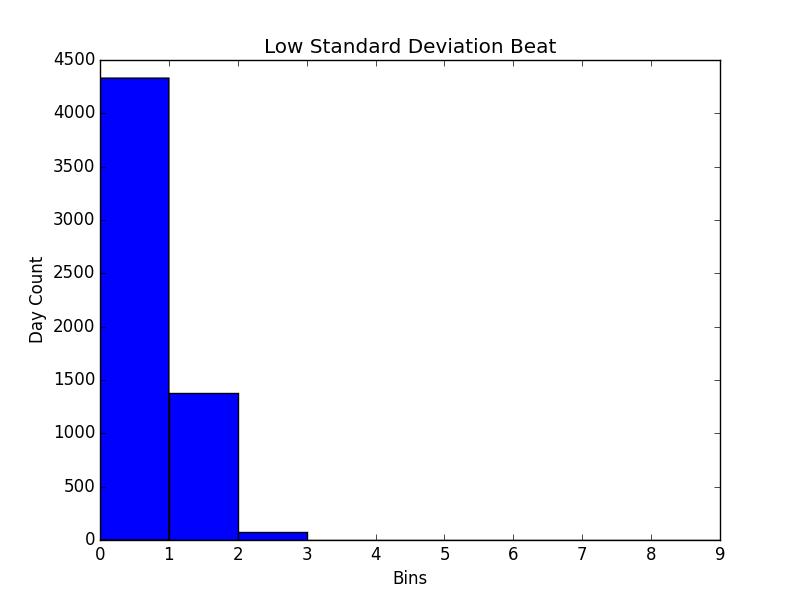}
\includegraphics[width=1.94in]{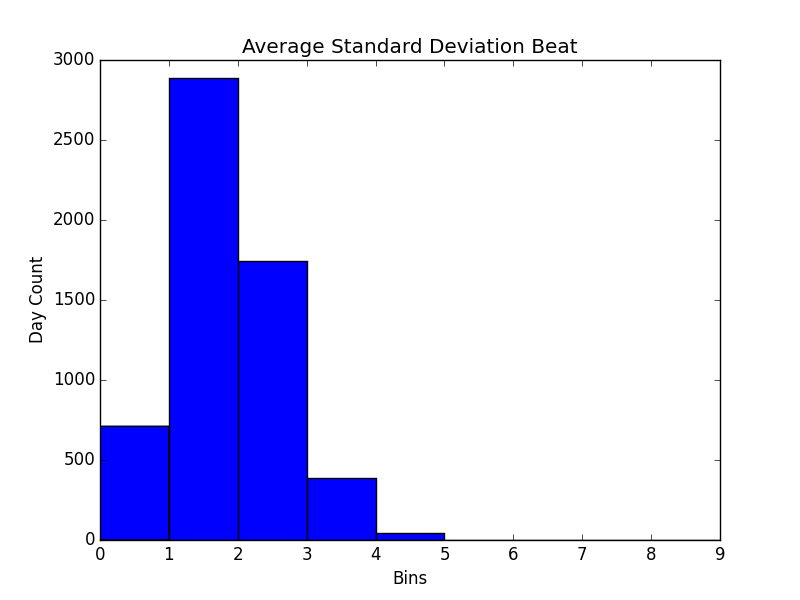}
\includegraphics[width=1.94in]{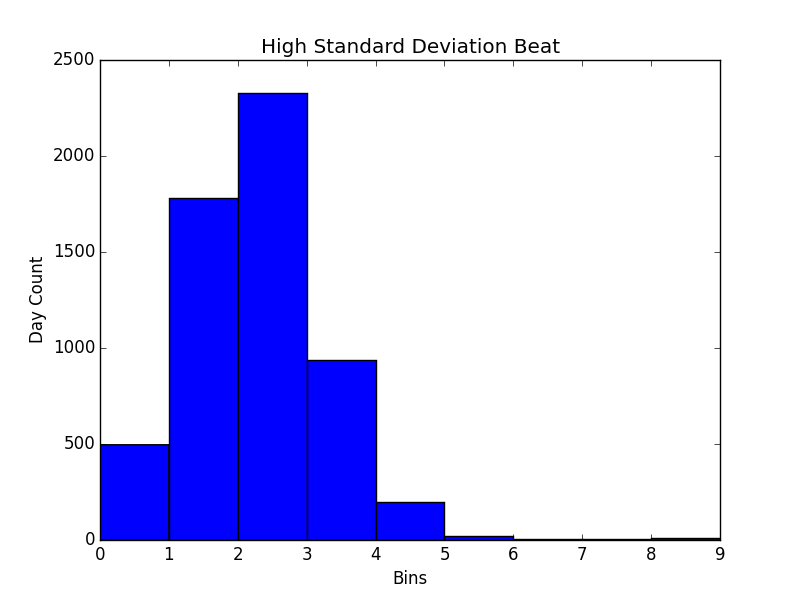}
\caption{Histograms of bin counts for three different beats in Chicago.}
\end{center}
\end{figure}

Aside from just aggregating the total count, we also obtain counts for other fields including different crime primary types (e.g. theft), number of arrests/nonarrests, and domestic crimes.
Including different primary types is important because they behave differently from the total count and from one another.
Figure 7 shows the different distributions for three different crime primary types, for which we use Assualt, Theft, and Narcotics type crimes.

For Portland, the dataset is not as rich, but it has several of the same fields.
The provided information does not include all reported crimes, but rather calls for service that the police received.
The calls are broken into groups based on type, and they include coordinates that allow for easy mapping.
Thus for Portland we are predicting the number of calls for service, rather than the raw crime count.
Unlike Chicago, the Portland data does not include any fields indicating whether or nor an arrest was made or if the crime was domestic, so these features are missing.

\begin{figure}[ht]
\begin{center}
\includegraphics[width=1.94in]{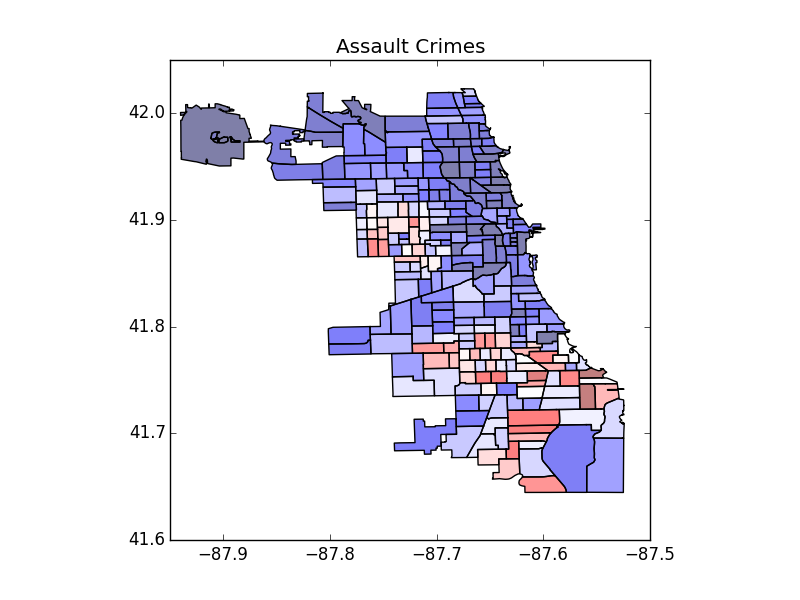}
\includegraphics[width=1.94in]{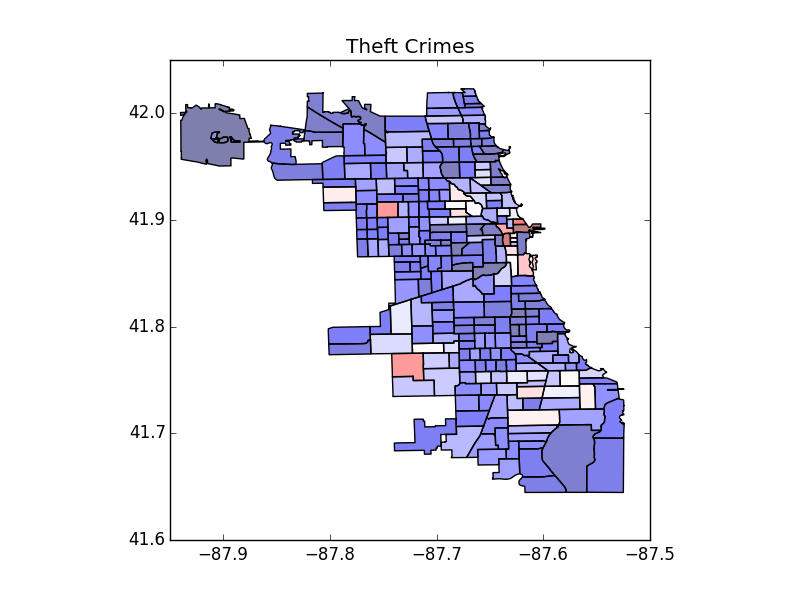}
\includegraphics[width=1.94in]{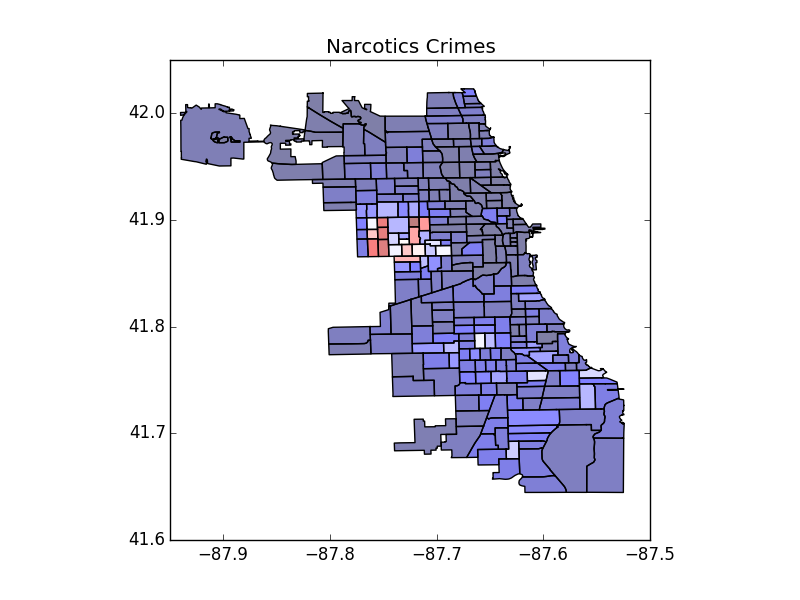}
\caption{Choropleth maps for counts of Assault, Theft, and Narcotics crime types in the Chicago beats over all records. Blue is lowest count and red is highest.}
\end{center}
\end{figure}

The public transportation data that is only used for Chicago, as no similar datasets are available for Portland.
The data provides station and line information for the Chicago Transit Authority (CTA) trains and buses.
Specifically, we use the daily ridership number for each train and bus line, as well as the number of entries at each train station.
Total riderships numbers are determined by summing up the entries at all stops along a given bus or train line.
This number is clearly not tied directly to the beats, but these numbers do affect the beats that each line passes through.
The latitude and longitude of each train station is available, and so each station is mapped to a beat, and then the number of entries is summed up for each beat.

The weather data is available at various weather stations across the country, but some of these stations have more complete records than others.
To map the data for each city, at each day we find the nearest available weather station with an entry in the relevant field and assign it to the beat or cell.
The NOAA provides a large amount of fields, but the majority are incomplete and so are not usable for our purposes.
The weather fields we use are temperature maximum, temperature minimum, midday temperature, precipitation, snowfall, and snow depth.

The census data is more static than the other datasets mentioned, and the only one that is not provided at a daily level.
We use data from the last two censuses, 2000 and 2010.
The census information we use for each day is the one that was taken closest to that day.
The census tracts are mapped into the beats and cells in a similar manner to the train stations from the transportation data.
There are too many features from this dataset to list here, but they include socioeconomic fields such as income, racial makeup, family information, age, etc..

For Chicago, we have the following feature breakdown. 
For FFN, for each beat there are approximately 40 crime specific features (e.g. number of arrests made and counts of crime sub-types), 6 weather features, 2 public transportation features, and around 40 census features.
About 10 of these crime features contain the history of the beat, such as the running average of the crime count for the past few days.
There are also non-beat features formed from the crime and transportation datasets.
For non-beat specific transportation features, we use the total ridership numbers for each bus and train line, of which there are approximately 200.
Additionally, this dataset includes day type information: including day of week, month of year, and day classification as a weekday, weekend, or holiday. 
For crime, we also include aggregation counts of different crime sub-types at the larger spatial region of police districts and community areas.
For models using a CNN, the top spatially correlated variables are used as channels.
The spatial correlation is calculated by determining the Global Moran Index of each feature, which takes into account both the feature's value and location.
Since having too many channels can greatly increase the computational cost, only the top eight spatially correlated variables are chosen for the CNN: the total count from the previous day, arrests made, domestic crimes, crimes without arrests, narcotics crimes, theft crimes, assault crimes, and CTA boardings.
For the RNN models, the features are essentially the same as the FFN models, except the beat's historical features are no longer needed since they are captured by the hidden state.
For Portland, the weather and census features are the same, and there is no transportation data set.
There are around 30 crime specific features per beat, and there are no non-beat features as with the Chicago data set.

\subsection{Training}
When training neural networks, typically the whole dataset is split into test and train sets, and then minibatches are randomly selected from the records in each set.
The model is trained on minibatches from the train set, and then its performance is evaluated on the test set.
For temporally static problems, it can be assumed that the parameters of the problem do not change, but this is not the case for crime.
It would make no sense evaluating on days in the past with a model trained on days that come after.
One way around this would be to have the training set be the first $~90\%$ of the data and the test set to be the remaining $~10\%$, but even this has problems.
It is highly likely that the optimal parameters for training records in 2001 have as much significance as those found in 2014 if evaluating on records in 2015.
Crime trends shift over time, and old records become less relevant as time goes on.
This means the time periods covered by the training sets should be smaller so they are more relevant to a specific evaluation period.

For this purpose, we use a walk-forward training method, and set the training sets to be 9 month periods, and the test sets to be the next 3 month period.
The process begins by training the model over the first 9 months of records.
After this training is complete, it is evaluated on the next 3 months of records.
The weights from this model are used as initial weights for the next model, which is trained on records over a shifted 9 month period, where the records from the previous test set make up the last 3 months of the new training set.
This process is continued step by step over the whole time period.
A snapshot of the process can be seen in Figure 8.
\begin{figure}[htbp]
\begin{center}
\includegraphics[width=3in]{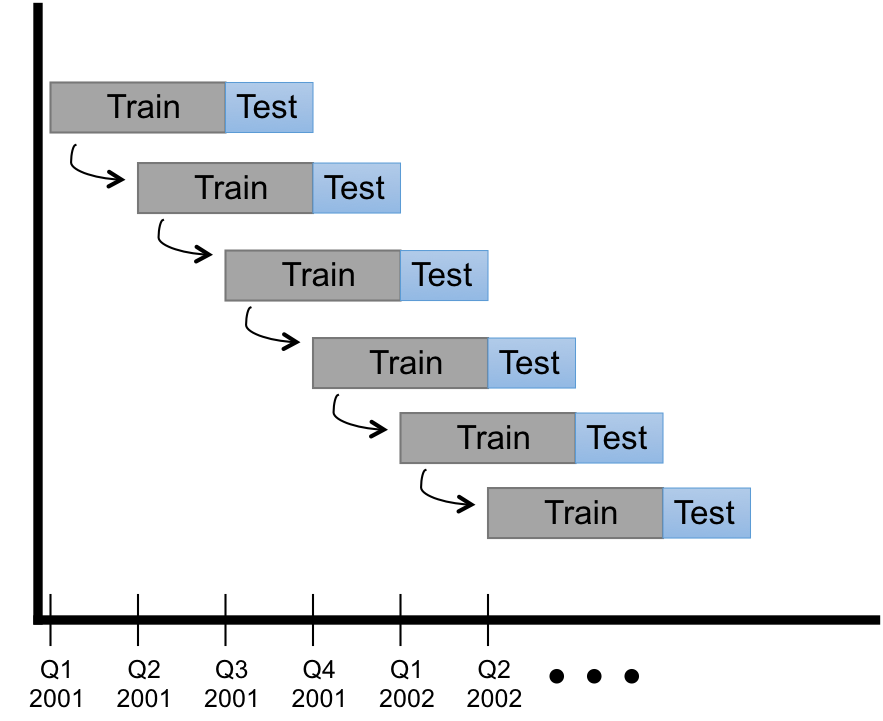}
\caption{Training process for walk-forward optimization.}
\label{default}
\end{center}
\end{figure}

\section{Experiments}
In this section, we discuss three experiments designed to test the accuracy and insights of our models.
The purpose of the first experiment is to establish the best network structure based off of the models described in Section 3.
The second experiment is about the importance of various datasets used.
The final experiment focuses on the importance of the features.

The first experiment uses all of the data sources described in Section 4 on each of the models described in Section 3.
In addition to the total crime count in each beat, we also ran experiments for three sub-types in Chicago.
Types 1, 2, and 3 correspond to violent crimes (e.g. assault), theft crimes, and narcotics crimes, respectively.
For the FFN, the model has two hidden layers and a softmax layer for each beat on top of these layers.
There are ten classes in each softmax layer corresponding to the following count bins: 0, 1-3, 4-6, 7-9, 10-12, 13-15, 16-20, 20-24, 25-29, 30+.
Similarly for the RNN, we use two layers, both of which are connected to the previous temporal time steps.

For the CNN models, we use an input grid of size 30 x 10.
For Portland we split the city into 300 cells so filling in the channels is simple, but for Chicago we map the beat coordinates onto a 30 x 10 grid and combine the features for beats falling in the same cell.
Since 30 x 10 is a small input for a CNN, we simply use the sequence of convolution, max pooling, and convolution.
This means the final output grid size is halved, 15 x 5, and the flattened output layer is of size 4800 because the last convolutional layer uses 64 filters.
After flattening, there is one hidden layer with half the dimensionality of the flattened layer.
This hidden layer is combined with the feed forward features.

The classification accuracies for each network and each prediction task is found in Table 1.
For each task, the FFN has the lowest accuracies and the RNN+CNN has the highest accuracies, which is expected.
The most common bin, corresponding to 1-3 crimes, occurs for 41\% of the records, and so these accuracies are well above this baseline.
Although there are engineered temporal and spatial features for the FFN, the network structure of the RNN+CNN is better able to account for these aspects of the problem with a simpler feature set.
This is likely because the engineered features have hard limits, such as average crime counts for the past four days or crime count on the previous day in the larger police districts, whereas the RNN+CNN model has no such restrictions.

In addition to classification accuracies, we also calculated the Mean Absolute Scaled Error (MASE) for each task using the top performing RNN+CNN model.
The MASE is calculated using the equation\begin{equation}
q_{t} = \frac{e_{t}}{\frac{1}{n-1}\sum_{t'=2}^{n}|Y_{t'}-Y_{t'-1}|} \nonumber
\end{equation}
where $e_{t}$ is the absolute value of the error at time $t$, $Y_{t'}$ is the actual value for the prediction at time $t$ and $n$ is the total number of predictions made.
We use MASE instead of the more popular MAPE since many actual values are $0$ leading to divisions by $0$ in MAPE.
Since we use bins instead of exact numerical predictions, the predicted values used are simply the midpoints of their bins (e.g. 2 for the 1-3 bin).
We calculate each $q_{t}$ for each beat or cell, and then take the average for each task.
An error less than one indicates the forecast is better than the one-step na\"{\i}ve prediction.
Using the best models for each task, we obtained the following MASE values: 0.74 for Chicago total, 0.77 for Chicago Type 1, 0.78 for Chicago Type 2, 0.83 for Chicago Type 3, and 0.77 for Portland Total.

\begin{table}[htp]
\begin{center}
{\small
\resizebox{\columnwidth}{!}{
\begin{tabular}{|l|c|c|c|c|c|}
\hline
 & Chicago Total & Chicago Type 1 & Chicago Type 2 & Chicago Type 3 & Portland Total \\
\hline
Feed Forward &  71.3 & 64.3 & 61.0 & 56.5 & 62.2\\
\hline
CNN & 72.7 & 65.1 & 62.7 & 56.9 &  62.9\\
\hline
RNN & 74.1 & 65.5 & 63.6 & 57.6 & 63.8 \\
\hline
RNN + CNN & \bf{75.6} & \bf{65.9} & \bf{64.7} & \bf{57.9} & \bf{65.3}\\
\hline
\end{tabular}
}
}
\end{center}
\caption{Classification accuracy results.}
\end{table}

Table 1 also shows that the different sub-types have different levels of improvement for the more complex models.
Type 2 crimes, corresponding to theft, have the highest improvement of all tasks when a CNN is added to the model.
On the other hand, Type 3 crimes, narcotics, have a significantly smaller improvement with the addition of only a CNN.
This suggests that spatial aspects are more important for predicting theft crimes than narcotics.
The Type 3 narcotics crimes did have a bigger improvement for models with an RNN, indicating that temporal patterns are more important for this type of crime.

For the second experiment, we use the RNN+CNN model for predicting Chicago total count by systematically removing individual datasets.
The size of these networks is the same as in the previous section, with the exception of a reduction in the input layer size.
In order to leverage the results of the previous experiment, we transfer over the weights from the input layer in the network corresponding to the remaining datasets as well as the full weights in the hidden layers.
We train using the same walk-forward method as in the first experiment.
The average classification accuracy decreases are reported in Table 2.

\begin{table}[htp]
\begin{center}
\begin{tabular}{|l|c|}
\hline
  & Accuracy Drop  \\
\hline
 Public Transportation &  2.3 \\
\hline
Weather &  0.7 \\
\hline
Census &  4.1 \\
\hline
\end{tabular}
\end{center}
\caption{Accuracy when excluding datasets from RNN+CNN model on total crime count for Chicago.}
\label{tab:dropout}
\end{table}

From Table 2, it is clear that although each dataset does not contribute equally, all of them do give some noticeable improvement to the classification accuracy.
Census data offering the largest improvement is not surprising, as there are well known links between socioeconomic factors and crime rates in communities. 
Public transportation also offers a significant improvement, likely because it gives a sense of the number of people and traffic through a given beat on given days.
With more people passing through a particular area of the city, there is an increased chance for crime.
Weather is the least informative of the three, since Chicago is a relatively small area and thus fairly homogeneous with respect to weather patterns, and as a result the values for weather do not vary much from beat to beat.

For the third and final experiment, we explore the conditions that affect model performance on the Chicago crime data.
For beat performance, the most accurate beats are those that have consistently low counts, e.g. falling in the first two bins.
However, this causes the model to generally under predict the actual count, since the model predicts the first two bins under most circumstances.
The second most accurate subset of beats are those with consistently high crime counts.
The predictions for the high count beats slightly trail the low count beats overall, but they are better at predictions when the actual count changes by one bin and much better when the actual count changes by two or more bins.
The mid count beats have the highest variability, with the performance values in between the high and low count beats.
These results are summarized in Table 3 for the Chicago total crime count on our RNN+CNN model.

\begin{table}[htp]
\begin{center}
\begin{tabular}{|l|c|c|c|}
\hline
  & Low & Mid & High  \\
\hline
Total accuracy &  79.3 &  72.3 &  77.4 \\
\hline
True count $\pm$ 1 bin &  74.4 &  70.1 &  75.5 \\
\hline
True count $\pm$ $>$2 bins &  55.3 & 58.4  &  64.7 \\
\hline
\end{tabular}
\end{center}
\caption{Accuracy for different subsets of beats. Low, Mid, and High correspond to the beats with the lowest, middle, and highest 10\% of crime counts.}
\label{tab:dropout}
\end{table}

Next, we study the impact of weather conditions on model accuracy.
To carry out this accuracy test we split the test records into ten percentile based groups for each weather condition.
The accuracies of these record subsets are plotted against the centers of the percentile bins in Figure 9.
We find that temperature does not have an effect on the accuracy any more than random noise around the overall accuracy.
Days with higher amounts of precipitation and snow, however, do decrease the accuracy of the best model slightly.
The bottom 65\% and 85\% of precipitation and snowfall values respectively are $0$, and so there is no significant difference in accuracy among these values.
However, looking at the days with the highest values of precipitation and snowfall, the accuracy does drop slightly.
It is unclear whether this is because these types of days are inherently less predictable or because there are fewer training examples for these days.
The effect of weather conditions on the accuracy from Figure 9 is consistent with the results of Experiment 2.

\begin{figure}[htbp]
\begin{center}
\includegraphics[width=10cm]{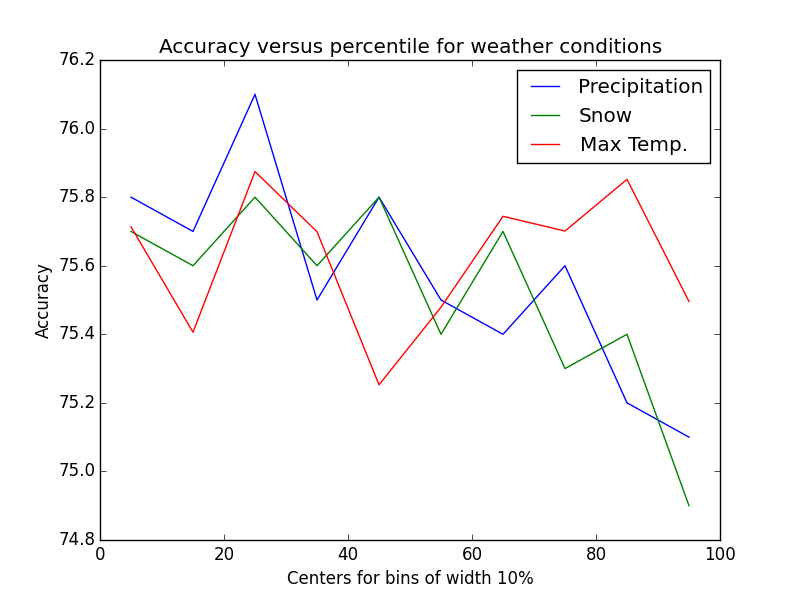}
\caption{Accuracies under different weather conditions}
\label{default}
\end{center}
\end{figure}

Finally, we look at the effect of the public transportation data.
While bus routes run across the whole city and through the vast majority of beats, the train lines run through a smaller subset of the beats.
An interesting result is that the accuracies of a beat containing a train station is on average $1.2\%$ higher than its neighboring beats.
The accuracy of a beat which contains one or more train lines passing through it (but not necessarily containing a station) is $0.5\%$ more accurate than its neighboring beats provided these beats do not also include a train line.
This indicates that both the station entries and total ridership of the line are important for those beats that contain a train line and/or station.
Also contained in the public transportation data is a flag for the type of day: weekday, weekend, and holiday.
Weekdays saw better performance than weekends and on average had an accuracy $0.9\%$ higher.
They also outperformed holidays by $1.1\%$.
This shows the model performs better during weekdays, which is not surprising because there are more events and activities that may happen on weekends or holidays, leading to less consistent conditions in the city.

\clearpage
\bibliographystyle{plain}
\bibliography{bib}

\end{document}